\documentclass{article} % For LaTeX2e
\usepackage{times}
\usepackage{hyperref}
\usepackage{url}

\usepackage{amsmath}
\usepackage{amssymb}
\usepackage{epsfig}
\usepackage{graphicx}
\usepackage[]{caption,subfig}
\usepackage{mdwlist}
\usepackage[lined, boxed]{algorithm2e}
\usepackage{multirow}

\graphicspath{{figs/}}

\title{Learning Robust Subspace Clustering}

\author{
Qiang Qiu~~~~~~ Guillermo Sapiro \\
Department of Electrical and Computer Engineering\\
Duke University\\
Durham, NC, 27708 \\
\texttt{\{qiang.qiu, guillermo.sapiro\}@duke.edu} \\
}

\begin{document}

\maketitle

\begin{abstract}
We propose a low-rank transformation-learning framework to robustify subspace clustering.
Many high-dimensional data, such as face images and motion sequences, lie in a union of low-dimensional subspaces.
The subspace clustering problem has been extensively studied in the literature to partition such high-dimensional data into clusters corresponding to their underlying low-dimensional subspaces.
However, low-dimensional intrinsic structures are often violated for real-world observations, as they can be corrupted by errors  or deviate from ideal models.
We propose to address this by learning a linear transformation on subspaces using matrix rank, via its convex surrogate nuclear norm, as the optimization criteria.
The learned linear transformation restores a low-rank structure for data from the same subspace, and, at the same time, forces a high-rank structure for data from different subspaces.
In this way, we reduce variations within the subspaces, and increase separations between the subspaces for more accurate subspace clustering.
This learned Robust Subspace Clustering  framework significantly enhances the performance of existing subspace clustering methods.
To exploit the low-rank structures of the transformed subspaces, we further introduce a subspace clustering technique, called Robust Sparse Subspace Clustering, which efficiently combines robust PCA with sparse modeling.
We also discuss the online learning of the transformation, and learning of the transformation while simultaneously reducing the data dimensionality.
Extensive experiments using public datasets are presented, showing that the proposed approach
significantly outperforms state-of-the-art subspace clustering methods.

\end{abstract}

\section{Introduction}

High-dimensional data often have a small intrinsic dimension.
For example, in the area of computer vision, face images of a subject \cite{9point}, \cite{Wright09}, handwritten images of a digit \cite{ocr}, and trajectories of a moving object \cite{sfm}, can all be well-approximated by a low-dimensional subspace of the high-dimensional ambient space. Thus, multiple class data often lie in a union of low-dimensional subspaces.
The \emph{subspace clustering} problem is to partition high-dimensional data into clusters corresponding to their underlying subspaces.

Standard clustering methods such as k-means in general are not applicable to subspace clustering. Various methods have been recently suggested for subspace clustering, such as Sparse Subspace Clustering (SSC) \cite{SSC} (see also its extensions and analysis in \cite{robustsubspace, ga-ssc, rssc, nssc}), Local Subspace Affinity (LSA) \cite{LSA}, Local Best-fit Flats (LBF) \cite{SLBF},
 Generalized Principal Component Analysis \cite{gpca},
 Agglomerative Lossy Compression \cite{alc},  Locally Linear Manifold Clustering \cite{llmc},
 and Spectral Curvature Clustering \cite{scc}. A recent survey on subspace clustering can be found in \cite{SubspaceClustering}.

Low-dimensional intrinsic structures, which enable subspace clustering,  are often violated for real-world computer vision observations (as well as other types of real data).
For example, under the assumption of Lambertian reflectance, \cite{9point} shows that face images of a subject obtained under a wide variety of lighting conditions can be approximated accurately with a 9-dimensional linear subspace. However,  real-world face images are often captured under pose variations; in addition, faces are not perfectly Lambertian, and exhibit cast shadows and specularities \cite{rpca}.
Therefore, it is critical for subspace clustering to handle corrupted underlying structures of realistic data, and as such, deviations from ideal subspaces.

When data from the same low-dimensional subspace are arranged as columns of a single matrix,  this matrix should be approximately low-rank.  Thus, a promising way to handle corrupted data for subspace clustering
is to restore such low-rank structure. Recent efforts have been invested in seeking transformations such that the transformed data can be decomposed as the sum of a low-rank matrix component and a sparse error one \cite{RASL, lrsalient, TILT}.
\cite{RASL} and \cite{TILT} are proposed for image alignment (see \cite{3dalign} for the extension to multiple-classes with applications in cryo-tomograhy), and \cite{lrsalient} is discussed in the context of salient object detection. All these methods build on recent theoretical and computational advances in rank minimization.

In this paper, we propose to robustify subspace clustering by learning a linear transformation on subspaces using matrix rank, via its nuclear norm convex surrogate,  as the optimization criteria.
The learned linear transformation recovers a low-rank structure for data from the same subspace, and, at the same time, forces a high-rank structure for data from different subspaces.
In this way, we reduce variations within the subspaces, and increase separations between the subspaces for more accurate subspace clustering.

This paper makes the following main contributions:
\begin{itemize*}
\item Subspace low-rank transformation is introduced in the context of subspace clustering;
\item A learned Robust Subspace Clustering framework is proposed to enhance existing subspace clustering methods;
\item We propose a specific subspace clustering technique, called Robust Sparse Subspace Clustering, by exploiting low-rank structures of the learned transformed subspaces;
\item We discuss online learning of subspace low-rank transformation for big data;
\item We discuss learning of subspace low-rank transformations with compression, where the learned matrix simultaneously reduces the data embedding dimension;
  \end{itemize*}

The proposed approach can be considered as a way of learning data features, with such features learned in order to reduce rank and encourage subspace clustering. As such, the framework and criteria here introduced can be incorporated into other data classification and clustering problems.

\section{Subspace Clustering using Low-rank Transformations}

Let $\{\mathcal{S}_c\}_{c=1}^C$ be $C$ $n$-dimensional subspaces of $\mathbb{R}^d$ (not all subspaces are necessarily of the same dimension, this is only here assumed to simplify notation).
Given a data set $\mathbf{Y} = \{ \mathbf{y}_i\}_{i=1}^N \subseteq \mathbb{R}^d$,
with each data point $\mathbf{y}_i$ in one of the $C$ subspaces, and in general the data arranged as columns of $Y$.
$\mathbf{Y}_c$ denotes the set of points in the $c$-th subspace $\mathcal{S}_c$,
and points are arranged as columns of the matrix $\mathbf{Y}_c$.
The \emph{subspace clustering} problem is to partition the data set $\mathbf{Y}$ into $C$ clusters corresponding to their underlying subspaces.

As data points in $\mathbf{Y}_c$ lie in a low-dimensional subspace, the matrix $\mathbf{Y}_c$ is expected to be \emph{low-rank}, and such low-rank structure is critical for accurate subspace clustering.
However, as discussed above, this low-rank structure is often violated for realistic data.

Our proposed approach is to robustify subspace clustering by learning a global linear transformation on
subspaces. Such linear transformation restores a low-rank structure for data from the same subspace, and, at the same time, encourages a high-rank structure for data from different subspaces. In this way, we reduce the variation within the subspaces and introduce separations between the subspaces for more accurate subspace clustering. In other words, the learned transform prepares the data for the ``ideal'' conditions of subspace clustering.

\subsection{Low-rank Transformation on Subspaces}

We now discuss low-rank transformation on subspaces in the context of subspace clustering.
We first introduce a method to learn a low-rank transformation using gradient descent (other optimization techniques could be considered). Then, we present
the online version for big data.
We further discuss the learning of a transformation with compression (dimensionality reduction) enabled.

\subsubsection{Problem Formulation}

We first assume the data cluster labels are known beforehand, assumption to be removed when discussing the full clustering approach in Section~\ref{sec:sc}.
We adopt matrix rank, actually its convex surrogate, as the key criterion, and compute one global linear transformation on all subspaces as
\begin{align} \label{stran_obj}
\underset{\mathbf{T}} \arg \min \frac{1}{C} \sum_{c=1}^C ||\mathbf{T Y}_c||_* - \lambda||\mathbf{T Y}||_*,
\end{align}
where $\mathbf{T} \in \mathbb{R}^{d \times d}$ is one global linear transformation on all data points,
and $||\mathbf{\cdot}||_*$ denotes the nuclear norm.
Intuitively, minimizing the first \emph{representation} term $\frac{1}{C} \sum_{c=1}^C ||\mathbf{T Y}_c||_*$ encourages a consistent representation for the transformed data from the same subspace; and minimizing the second \emph{discrimination} term $-||\mathbf{T Y}||_*$ encourages a diverse representation for transformed data from different subspaces. The parameter $\lambda \ge 0$  balances between the representation and discrimination.

\subsubsection{Gradient Descent Learning}

Given any matrix $\mathbf{A}$ of rank at most $r$,  the matrix norm $||\mathbf{A}||$ is equal to its largest singular value, and the nuclear norm $||\mathbf{A}||_*$ is equal to the sum of its singular values. Thus, these two norms are related by the inequality,
\begin{align} \label{norm_ineq}
||\mathbf{A}|| \le ||\mathbf{A}||_* \le r||\mathbf{A}||.
\end{align}

We use a simple gradient descent (though other modern nuclear norm optimization techniques could be considered, including recent real-time formulations \cite{pablo-lr}) to search for the transformation matrix $\mathbf{T}$ that minimizes (\ref{stran_obj}). The partial derivative of (\ref{stran_obj}) w.r.t $\mathbf{T}$ is written as,
\begin{align} \label{stran_der}
\frac{\partial}{\partial \mathbf{T}}[  \frac{1}{C} \sum_{c=1}^C ||\mathbf{T Y}_c||_* - \lambda||\mathbf{T Y}||_* ].
\end{align}

Due to property (\ref{norm_ineq}), by minimizing the matrix norm, one also minimizes an upper bound to the nuclear norm. (\ref{stran_der}) can now be evaluated as,
\begin{align} \label{stran_sub}
\Delta \mathbf{T} =  \frac{1}{C} \sum_{c=1}^C \partial ||\mathbf{T Y}_c||\mathbf{Y}_c^T - \lambda \partial ||\mathbf{T Y}||\mathbf{Y}^T,
\end{align}
where $\partial ||\mathbf{\cdot}||$ is the subdifferential of norm $||\mathbf{\cdot}||$. Given a matrix $\mathbf{A}$, the subdifferential $\partial ||\mathbf{A}||$ can be evaluated using the simple approach shown in Algorithm~\ref{subdifferential} \cite{subdifferential}. By evaluating $\Delta \mathbf{T}$, the optimal transformation matrix $\mathbf{T}$ can be searched with gradient descent
$
\mathbf{T}^{(t+1)} = \mathbf{T}^{(t)} - \nu \Delta \mathbf{T} ,
$
where $\nu > 0$ defines the step size.
After each iteration, we normalize $\mathbf{T}$ as $\frac{\mathbf{T}}{||\mathbf{T}||}$. This algorithm guarantees convergence to a local minimum.

\begin{algorithm}[ht]
%\SetAlFnt{\footnotesize \sf}
\footnotesize
\KwIn{An $m \times n$ matrix $\mathbf{A}$, a small threshold value $\delta$}
\KwOut{The subdifferential of the matrix norm $\partial ||\mathbf{A}||$.}
\Begin{
\BlankLine
1. Perform singular value decomposition: \\
$\mathbf{A}=\mathbf{U} \mathbf{\Sigma} \mathbf{V}$ \;
\BlankLine
2. $s \leftarrow$ the number of singular values smaller than $\delta$ , \\
3. Partition $\mathbf{U}$ and $\mathbf{V}$ as \\
$\mathbf{U} = [\mathbf{U}^{(1)}, \mathbf{U}^{(2)}]$, $\mathbf{V} = [\mathbf{V}^{(1)}, \mathbf{V}^{(2)}]$ \;
where $\mathbf{U}^{(1)}$ and $\mathbf{V}^{(1)}$ have $(n-s)$ columns. \\
\BlankLine
4. Generate a random matrix $\mathbf{B}$ of the size $(m-n+s)\times s$, \\
$\mathbf{B} \leftarrow \frac{\mathbf{B}}{||\mathbf{B}||}$ \;
\BlankLine
5. $\partial ||\mathbf{A}|| \leftarrow \mathbf{U}^{(1)} \mathbf{V}^{(1)T} + \mathbf{U}^{(2)} \mathbf{B} \mathbf{V}^{(2)T}$ \;
\BlankLine
6. Return $\partial ||\mathbf{A}||$ \;

}
\caption{An approach to evaluate the subdifferential of a matrix norm.}
\label{subdifferential}
\end{algorithm}

\subsubsection{Online Learning}

When data $\mathbf{Y}$ is big, we use an online algorithm to learn the low-rank transformation on subspaces:

\begin{itemize*}
\item We first randomly partition the data set $\mathbf{Y}$ into $B$ mini-batches;
\item Using mini-batch gradient descent, a variant of stochastic gradient descent, the gradient $\Delta \mathbf{T}$ is approximated by a sum of gradients obtained from each mini-batch of samples,
$\mathbf{T}^{(t+1)} = \mathbf{T}^{(t)} - \nu \sum_{b=1}^B \Delta \mathbf{T}_b$,
%\begin{align} \label{olgradstep}
%\mathbf{T}^{(t+1)} = \mathbf{T}^{(t)} - \nu \sum_{b=1}^B \Delta \mathbf{T}_b,
%\end{align}
%\begin{align} \label{olgradstep}
% \Delta \mathbf{T} =  \sum_{b=1}^B \Delta \mathbf{T}_b,
%\end{align}
where $\Delta \mathbf{T}_b$ is obtained from (\ref{stran_sub}) using only data points in the $b$-th mini-batch;
\item Starting with the first mini-batch, we learn subspace transformation $\mathbf{T}_b$ using data only in the $b$-th mini-batch, with $\mathbf{T}_{b-1}$ as warm restart.
\end{itemize*}

\subsubsection{Subspace Transformation with Compression}

Given data $\mathbf{Y} \subseteq \mathbb{R}^d$, so far, we considered a square linear transformation $\mathbf{T}$ of size $d \times d$. If we devise a ``fat" linear transformation $\mathbf{T}$ of size $r \times d$, where $(r<d)$, we enable dimension reduction along with transformation (the above discussed algorithm is directly applicable to learning a linear transformation with less rows than columns).
This connects the proposed framework with the literature on compressed sensing, though the goal here is to learn a sensing matrix $\mathbf{T}$ for subspace classification and not for reconstruction \cite{CS1}. The nuclear-norm minimization provides a new metric for such sensing design paradigm.

\subsection{Learning for Subspace Clustering}
\label{sec:sc}

We now first present a general procedure to enhance the performance of existing subspace clustering methods in the literature. Then we further propose a specific subspace clustering technique to fully exploit the low-rank structure of (learned) transformed subspaces.

\subsubsection{A Learned Robust Subspace Clustering (RSC) Framework}

The data labeling (clustering)  is not known beforehand in practice.  The proposed algorithm, Algorithm~\ref{algorsc}, iterates between two stages:
In the first assignment stage, we obtain clusters using any subspace clustering methods, e.g., SSC \cite{SSC}, LSA \cite{LSA}, LBF \cite{SLBF}.
In particular, in this paper we often use the new technique introduced in Section~\ref{sec:r-ssc}.
In the second update stage, based on the current clustering result, we compute the optimal subspace transformation that minimizes (\ref{stran_obj}). The algorithm is repeated until the clustering assignments stop changing. Algorithm~\ref{algorsc} is a general procedure to enhance the performance of any subspace clustering methods.

\begin{algorithm}[ht]
%\SetAlFnt{\footnotesize \sf}
\footnotesize
\KwIn{A set of data points $\mathbf{Y} = \{ \mathbf{y}_i\}_{i=1}^N \subseteq \mathbb{R}^d$ in a union of $C$ subspaces.}
\KwOut{A partition of $\mathbf{Y}$ into $C$ disjoint clusters $\{ \mathbf{Y}_c\}_{c=1}^C$ based on underlying subspaces.}
\Begin{
\BlankLine
1. Initial a transformation matrix $\mathbf{T}$ as the identity matrix \;
\BlankLine
\Repeat {assignment convergence}{
\textbf{{Assignment stage:}}\\
2. Assign points in $\mathbf{TY}$  to clusters with any subspace clustering methods, e.g., the proposed R-SSC\;
\BlankLine
\textbf{{Update stage:}}\\
3. Obtain transformation $\mathbf{T}$ by minimizing (\ref{stran_obj}) based on the current clustering result \;
}
\BlankLine
4. Return the current clustering result $\{ \mathbf{Y}_c\}_{c=1}^C$ \;
}
\caption{Learning a robust subspace clustering framework.}
\label{algorsc}
\end{algorithm}

\subsubsection{Robust Sparse Subspace Clustering (R-SSC)}
\label{sec:r-ssc}

Though Algorithm~\ref{algorsc} can adopt any subspace clustering methods, to fully exploit the low-rank structure of transformed subspaces, we further propose the following specific technique for the clustering step in the RSC framework, called Robust Sparse Subspace Clustering (R-SSC):

\begin{itemize*}
\item For the transformed subspaces, we first recover their low-rank representation $\mathbf{L}$ by performing a low-rank decomposition (\ref{rpca}), e.g., using RPCA \cite{rpca},
    \begin{align} \label{rpca}
\underset{\mathbf{L}, \mathbf{S}} \arg \min ||\mathbf{L}||_* + \beta ||\mathbf{S}||_1 ~~s.t.~ \mathbf{TY} =\mathbf{L}+\mathbf{S}.
\end{align}

\item Each transformed point $\mathbf{Ty}_i$ is then sparsely decomposed over $\mathbf{L}$,
\begin{align} \label{SSC}
\underset{\mathbf{x}_i} \arg \min \|\mathbf{Ty}_i-\mathbf{L}\mathbf{x}_i\|_{2}^{2} ~~s.t.~ \|\mathbf{x}_i\|_{0}\leq K ,
\end{align}
where $K$ is a predefined sparsity value ($K > d$).
As explained in \cite{SSC}, a data point in a linear or affine subspace of dimension $d$ can be written as a linear or affine combination of $d$ or $d+1$ points in the same subspace. Thus, if we represent a point as a linear or affine combination of all other points, a sparse linear or affine combination can be obtained by choosing $d$ or $d+1$ nonzero coefficients.

\item As the optimization process for (\ref{SSC}) is computationally demanding, we further simplify (\ref{SSC}) using Local Linear Embedding \cite{LLE}, \cite{llc}. Each transformed point $\mathbf{Ty}_i$ is represented using its $K$ Nearest Neighbors (NN) in $\mathbf{L}$, which are denoted as $\mathbf{L}_i$,
 \begin{align} \label{LLE}
 \underset{\mathbf{x}_i} \arg \min \|\mathbf{Ty}_i-\mathbf{L}_i\mathbf{x}_i\|_{2}^{2} ~~s.t.~ \|\mathbf{x}_i\|_{1} = 1 .
\end{align}
Let $\mathbf{L}_i$ be $\mathbf{Ty}_i$ centered through $\mathbf{\bar{L}}_i = \mathbf{L}_i- \mathbf{1}\mathbf{Ty}_i^T$. $\mathbf{x}_i$ can then  be obtained in closed form,
\[
\mathbf{x}_i = \mathbf{\bar{L}}_i \mathbf{\bar{L}}_i^T \setminus \mathbf{1} ,
\]
where $\mathbf{x} = \mathbf{A} \setminus \mathbf{B}$ solves the system of linear equations $\mathbf{A}\mathbf{x} = \mathbf{B}$.
As suggested in \cite{LLE}, if the correlation matrix $\mathbf{\bar{L}}_i \mathbf{\bar{L}}_i^T$ is nearly singular, it can be conditioned by adding a small multiple of the identity matrix.

\item Given the sparse representation $\mathbf{x}_i$ of each transformed data point $\mathbf{Ty}_i$, we denote the sparse representation matrix as $\mathbf{X} = [\mathbf{x}_1 \ldots \mathbf{x}_N]$. It is noted that $\mathbf{x}_i$ is written as an $N$-sized vector with no more than $K$ non-zero values ($N$ being the total number of data points).
    The pairwise affinity matrix is now defined as
    $
    \mathbf{W} = |\mathbf{X}|+|\mathbf{X}^T|,
    $
and the subspace clustering is obtained using spectral clustering \cite{spectral}.
\end{itemize*}

Based on experimental results presented in Section~\ref{sec:exp}, the proposed R-SSC outperforms  state-of-the-art subspace clustering techniques, both in accuracy and running time, e.g., about 500 times faster than the original SSC using the implementation provided in \cite{SSC}.  The accuracy is even further enhanced when R-SCC is used as an internal step of RSC.

\section{Experimental Evaluation}
\label{sec:exp}

This section presents experimental evaluations on three public datasets (standard benchmarks): the MNIST handwritten digit dataset, the Extended YaleB face dataset \cite{yaleb}, and the CMU Motion Capture (Mocap) dataset at \url{http://mocap.cs.cmu.edu}.
 The MNIST dataset consists of 8-bit grayscale handwritten digit images of ``0"  through
``9" and 7000 examples for each class.
The Extended YaleB face dataset contains 38 subjects with near frontal pose under 64 lighting conditions.
All the images are resized to $16 \times 16$.
The Mocap dataset contains measurements of 42 (non-imaging) sensors that capture the motions of 149 subjects performing multiple actions.

Subspace clustering methods compared are SSC \cite{SSC}, LSA \cite{LSA}, and LBF \cite{SLBF}.
Based on the studies in \cite{SSC}, \cite{SubspaceClustering} and \cite{SLBF}, these three methods exhibit state-of-the-art  subspace clustering performance.
We adopt the LSA and SSC implementations provided in \cite{SSC} from \url{http://www.vision.jhu.edu/code/}, and the LBF implementation provided in \cite{SLBF} from
\url{http://www.ima.umn.edu/~zhang620/lbf/}.

\subsection{Evaluation with Illustrative Examples}

\begin{figure*} [ht]
\centering
  \subfloat[Original subspaces for digits \{1, 2\}.] {\label{fig:subcode2_profile} \includegraphics[angle=0, height=0.16\textwidth, width=.5\textwidth]{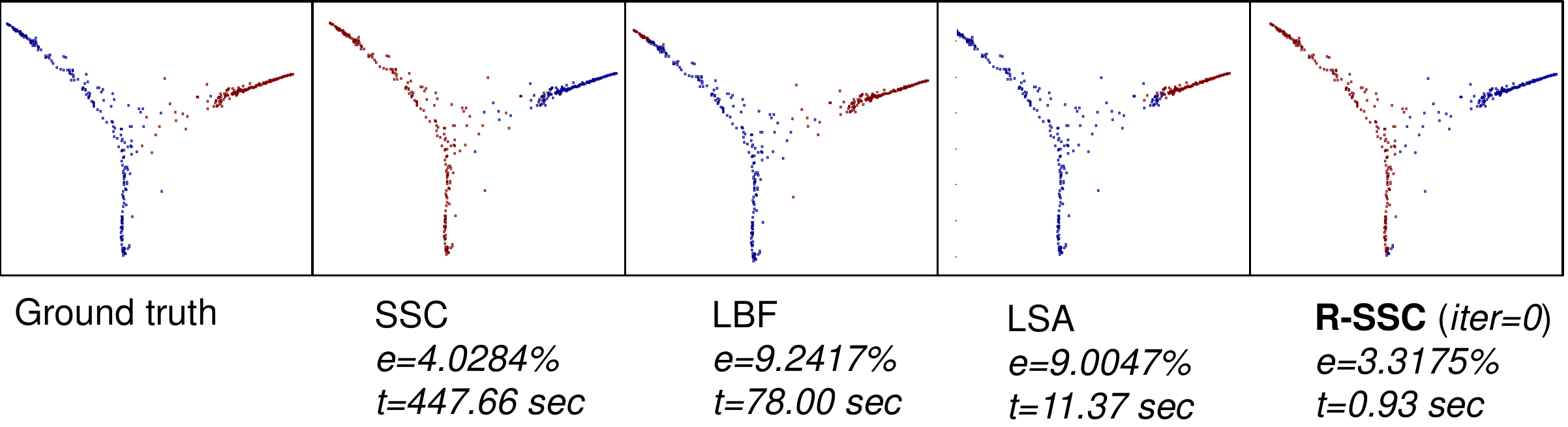}}
 \subfloat[Transformed subspaces for digits \{1, 2\}.] {\label{fig:TF_subcode2_front} \includegraphics[angle=0, height=0.18\textwidth, width=.5\textwidth]{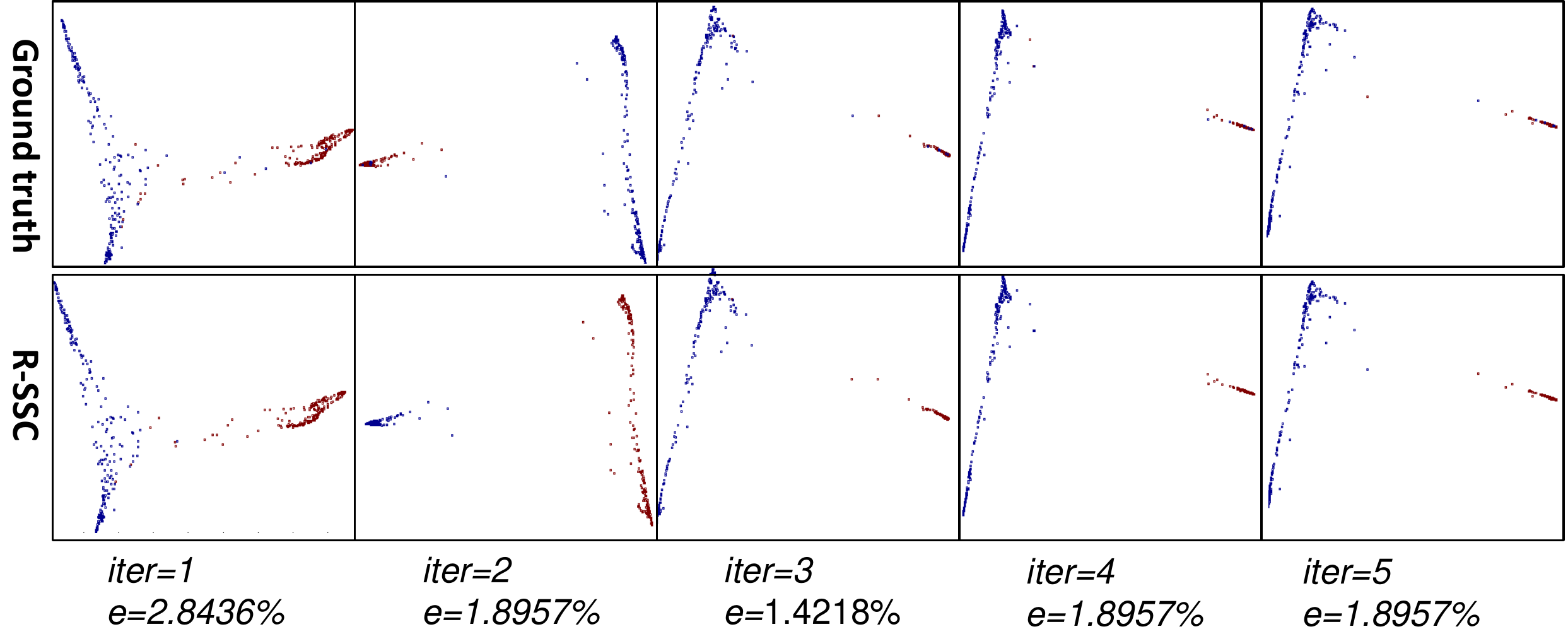} } \\
 \subfloat[Original subspaces for digits \{1, 7\}.] {\label{fig:TF_subcode2_side} \includegraphics[angle=0, height=0.16\textwidth, width=.5\textwidth]{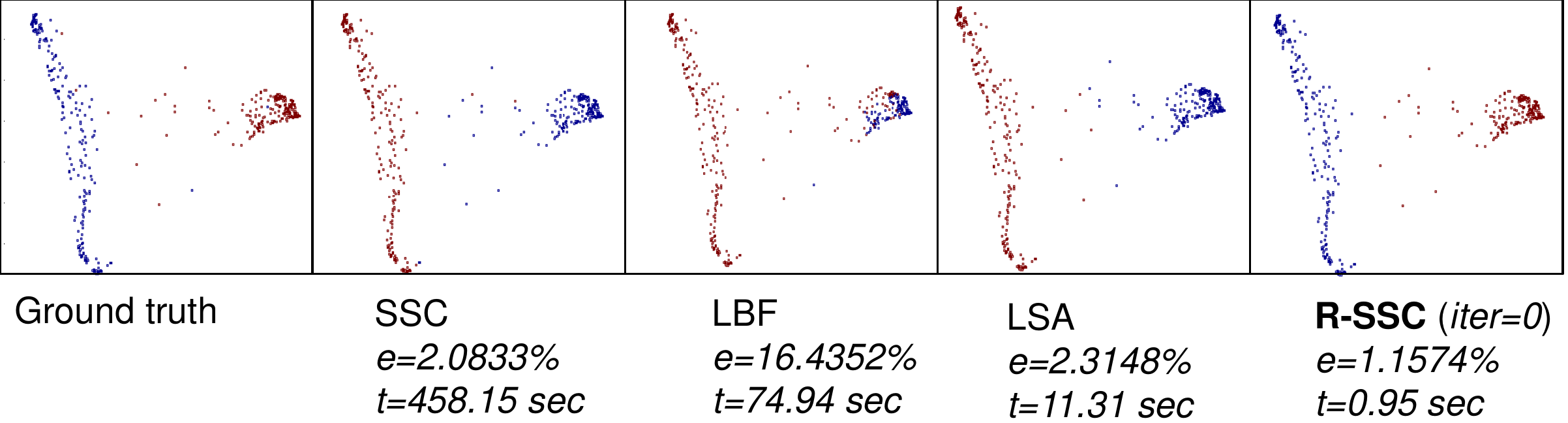}}
  \subfloat[Transformed subspaces for digits \{1, 7\}.] {\label{fig:TF_subcode2_profile} \includegraphics[angle=0, height=0.18\textwidth, width=.5\textwidth]{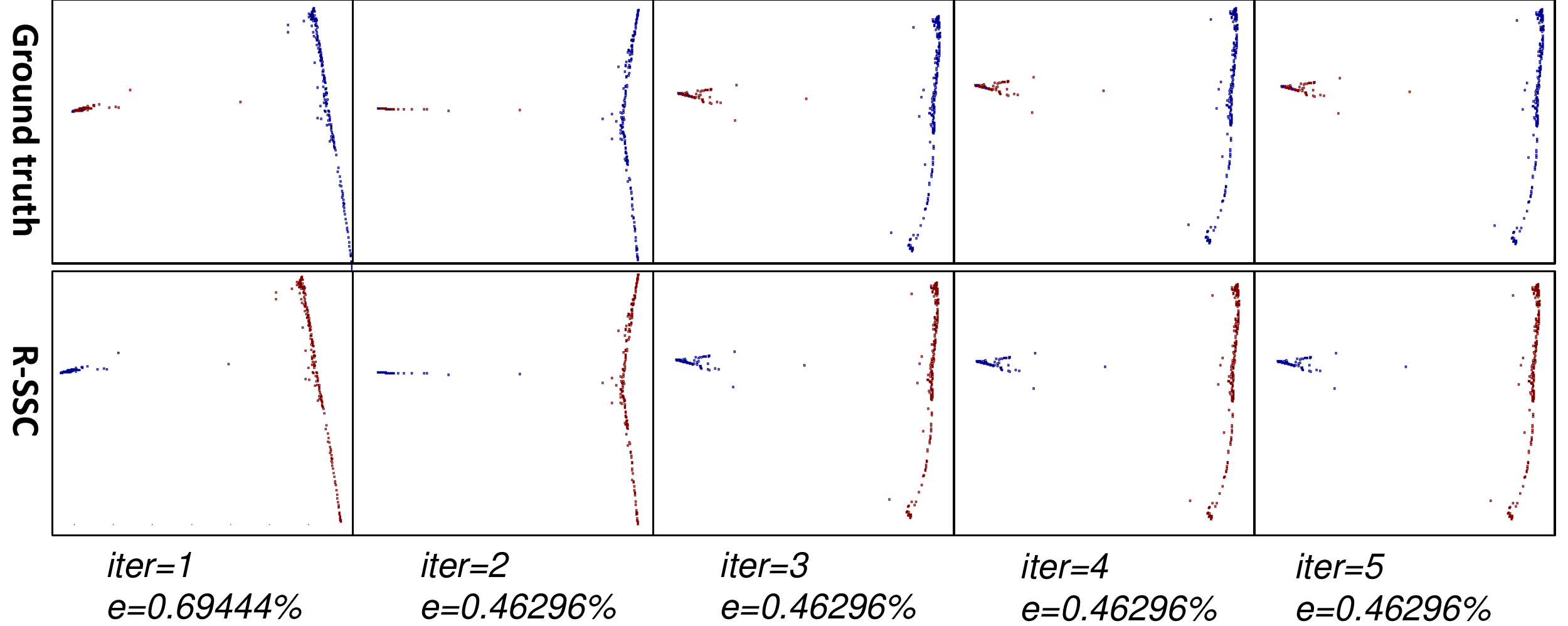}} \\
\caption{Misclassification rate (\emph{e})  and running time (\emph{t}) on clustering 2 digits. Methods compared are SSC \cite{SSC}, LSA \cite{LSA}, and LBF \cite{SLBF}.
For visualization, the data are plotted with the dimension reduced to 2 using Laplacian Eigenmaps \cite{eigenmap}.
Different clusters are represented by different colors and the \emph{ground truth} is plotted with the true cluster labels.
$iter$ indicates the number of RSC iterations in Algorithm~\ref{algorsc}.
The proposed R-SSC outperforms state-of-the-art methods in terms of both clustering accuracy and running time, e.g., about 500 times faster than SSC.
The clustering performance of R-SSC is further improved using the proposed RSC framework.
Note how the data clearly cluster in clean subspaces in the transformed domain (best viewed zooming on screen).
}
\label{fig:2digit}
%\end{center}
\end{figure*}

\begin{figure*} [ht]
\centering
 \subfloat[Digits \{1, 2, 3\}.] {\label{fig:subcode2_front} \includegraphics[angle=0, height=0.15\textwidth, width=.48\textwidth]{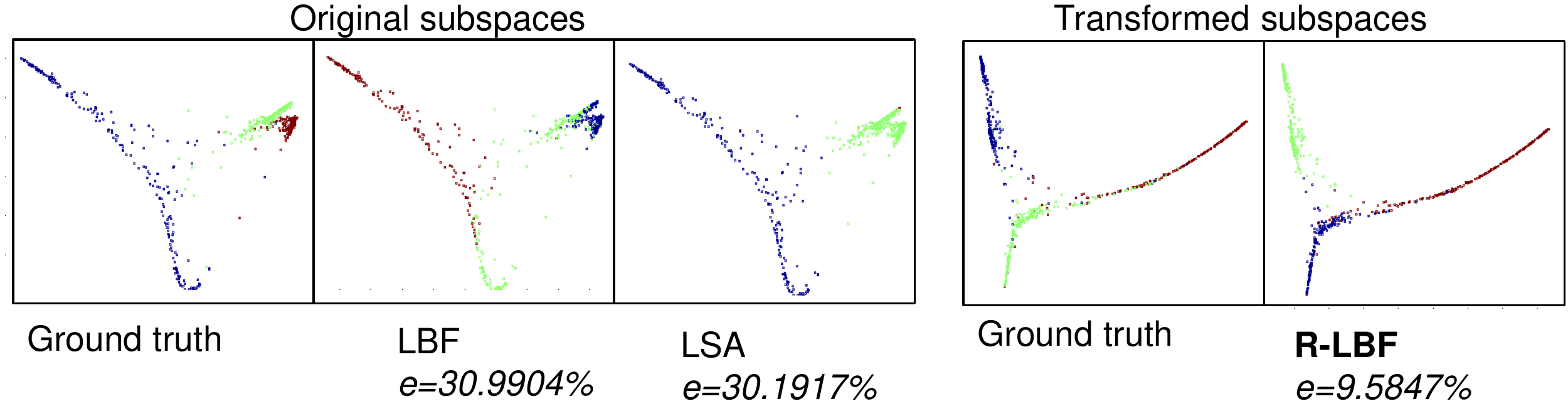} \hspace{10pt}}
  \subfloat[Digits \{2, 4, 8\}.] {\label{fig:subcode2_profile} \includegraphics[angle=0, height=0.15\textwidth, width=.48\textwidth]{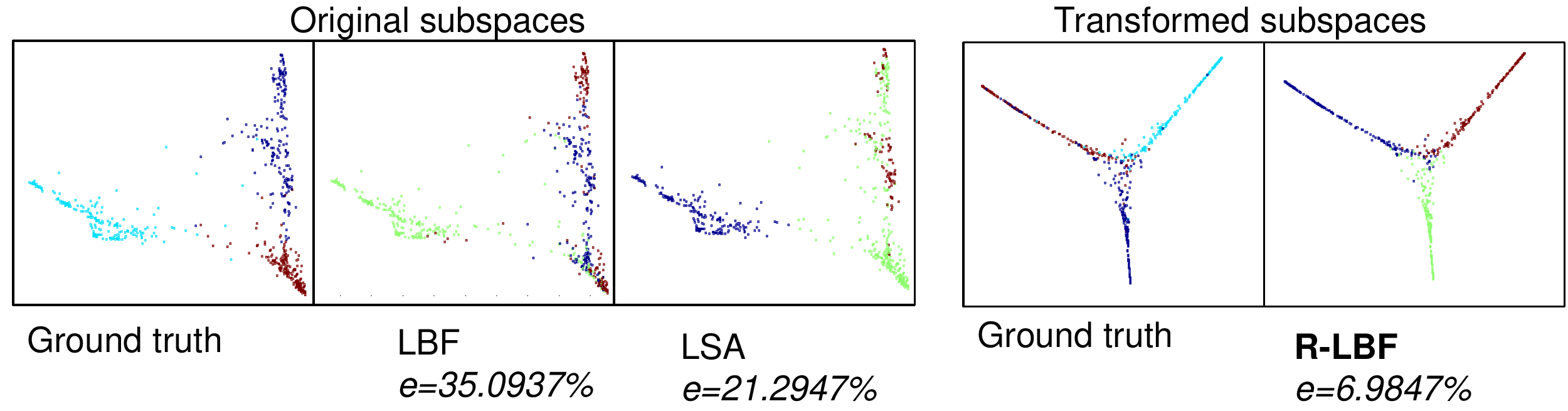}}
\caption{Misclassification rate (\emph{e}) on clustering 3 digits.
Methods compared are LSA \cite{LSA} and LBF \cite{SLBF}.
LBF is adopted in the proposed RSC framework and denoted as {R-LBF}.
After convergence, R-LBF significantly outperforms state-of-the-art methods.}
\label{fig:3digit}
%\end{center}
\end{figure*}

We conduct the first set of experiments on a subset of the MNIST dataset.
We adopt a similar setup as described in \cite{SLBF}, using the same sets of 2 or 3 digits, and randomly choose 200 images for each digit. We do not perform dimension reduction to preprocess the data as \cite{SLBF}. We set the sparsity value $K=6$ for R-SSC, and perform $100$ iterations for the gradient descent updates while learning the transformation on subspaces.

Fig.~\ref{fig:2digit} shows the misclassification rate (\emph{e})  and running time (\emph{t}) on clustering subspaces of two digits.
The misclassification rate is the ratio of misclassified points to the total number of points.
For visualization  purposes, the data are plotted with the dimension reduced to 2 using Laplacian Eigenmaps \cite{eigenmap}.
Different clusters are represented by different colors and the ground truth is plotted using the true cluster labels.
The proposed R-SSC outperforms state-of-the-art methods, both in terms of clustering accuracy and running time.
The clustering error of R-SSC is further reduced using the proposed RSC framework in Algorithm~\ref{algorsc} through the learned low-rank subspace transformation.
The clustering results converge after about 3 RSC iterations.
After convergence, the learned subspace transformation not only recovers a low-rank structure for
data from the same subspace, but also increases the separations between the subspaces for more accurate clustering.

Fig.~\ref{fig:3digit} shows misclassification rate (\emph{e}) on clustering subspaces of three digits.
Here we adopt LBF in our RSC framework, denoted as Robust LBF (R-LBF),  to illustrate that the performance of existing subspace clustering methods can be enhanced using the proposed RSC framework.
After convergence, R-LBF, which uses the proposed learned subspace transformation, significantly outperforms state-of-the-art methods.

\subsubsection{Online vs. Batch Learning}

\begin{figure*} [ht]
\centering
 \subfloat[Batch learning with various $\lambda$ values.] {\label{fig:batchconv} \includegraphics[angle=0, height=0.17\textwidth, width=.32\textwidth]{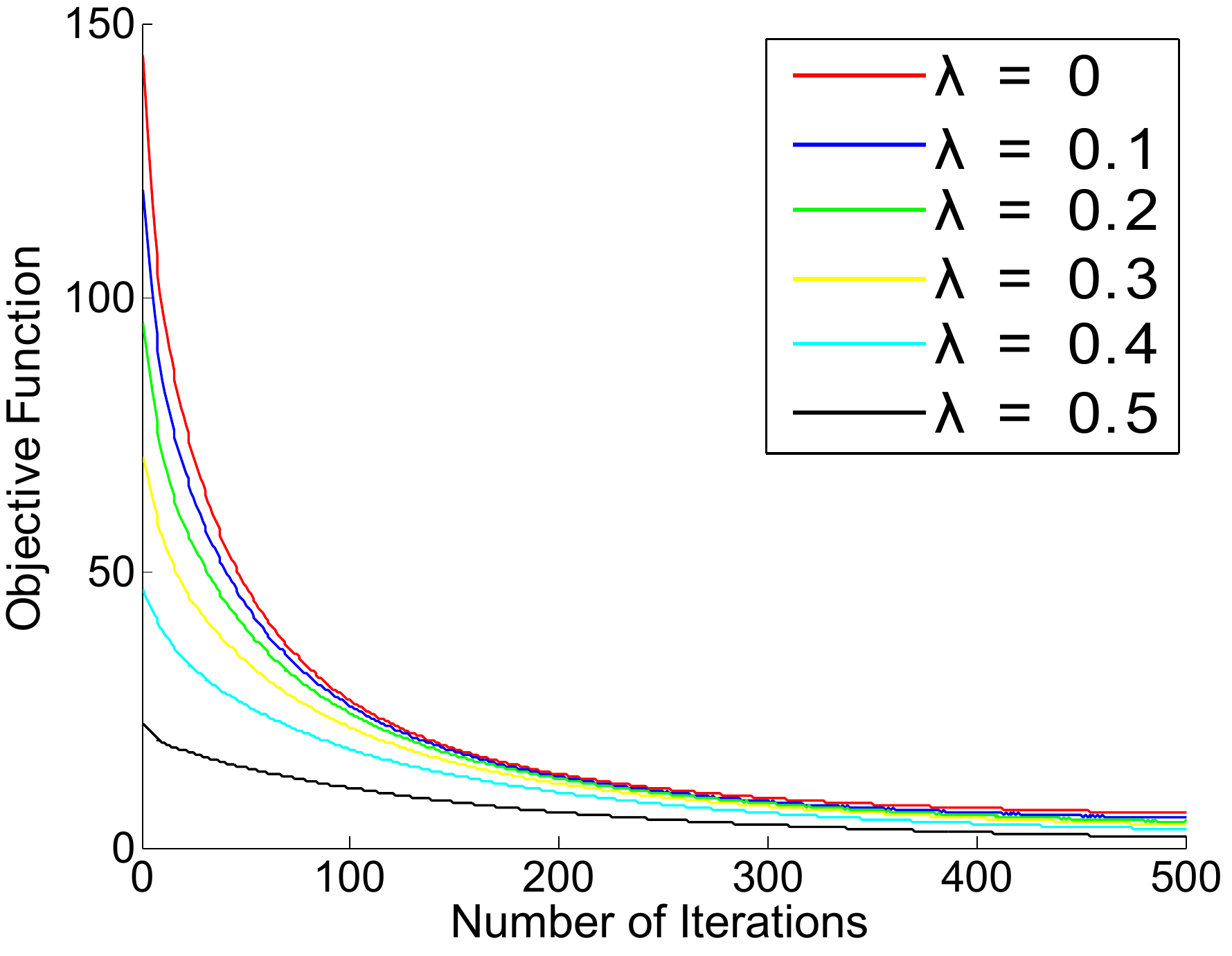} \hspace{20pt}}
  \subfloat[Online vs. batch learning ($\lambda=0.5$).] {\label{fig:olconv} \includegraphics[angle=0, height=0.17\textwidth, width=.4\textwidth]{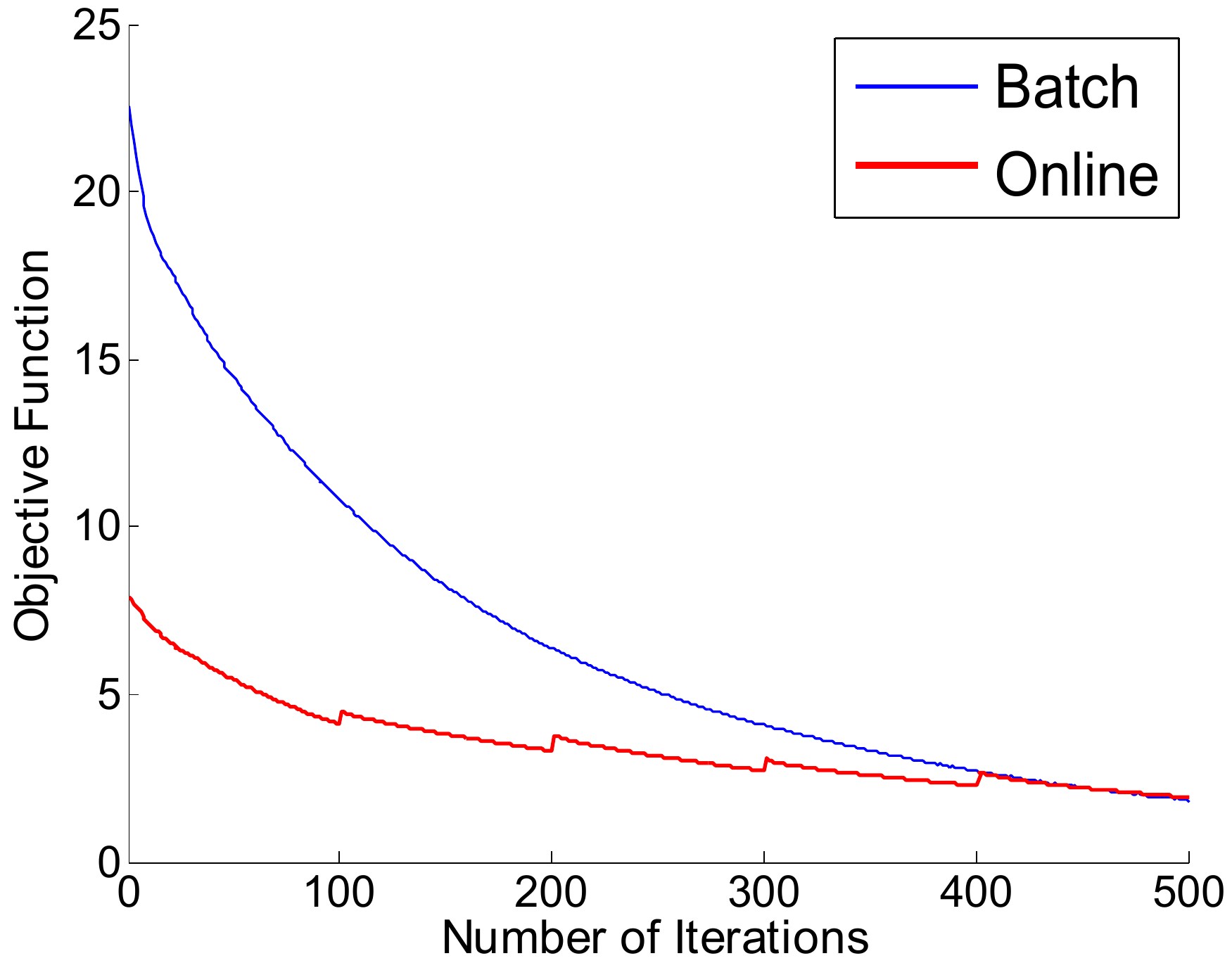}}
\caption{Convergence of the objective function (\ref{stran_obj}) using online and batch learning for subspace transformation. We always observe empirical convergence for both online and batch learning.
In (b), to converge to the same objective function value, it takes $131.76$ sec. for online learning and $700.27$ sec. for batch learning.
}
\label{fig:convergence}
%\end{center}
\end{figure*}

In this set of experiments, we use digits \{1, 2\} from the MNIST dataset. We select 1000 images for each digit, and randomly partition them into 5 mini-batches.
We first perform one iteration of RSC in Algorithm~\ref{algorsc} over all selected data with various $\lambda$ values.
 As shown in Fig.~\ref{fig:batchconv}, we always observe empirical convergence for subspace transformation learning in (\ref{stran_obj}).

As discussed, the value of $\lambda$ balances between the representation and discrimination terms in the objective function (\ref{stran_obj}). In general, the value of $\lambda$ can be estimated through cross-validations. In our experiments, we always choose $\lambda=\frac{1}{C}$, where $C$ is the number of subspaces.

Starting with the first mini-batch, we then perform one iteration of RSC  over one mini-batch a time, with the subspace transformation learned from the previous mini-batch as warm restart. We adopt here $100$ iterations for the gradient descent updates. As shown in Fig.~\ref{fig:olconv}, we observe similar empirical convergence for online transformation learning. To converge to the same objective function value, it takes
$131.76$ sec. for online learning and $700.27$ sec. for batch learning.

\subsection{Application to Face Clustering}

\begin{figure*} [ht]
\centering
 \subfloat[Example illumination conditions.] {\label{fig:yalelight} \includegraphics[angle=0, height=0.07\textwidth, width=.5\textwidth]{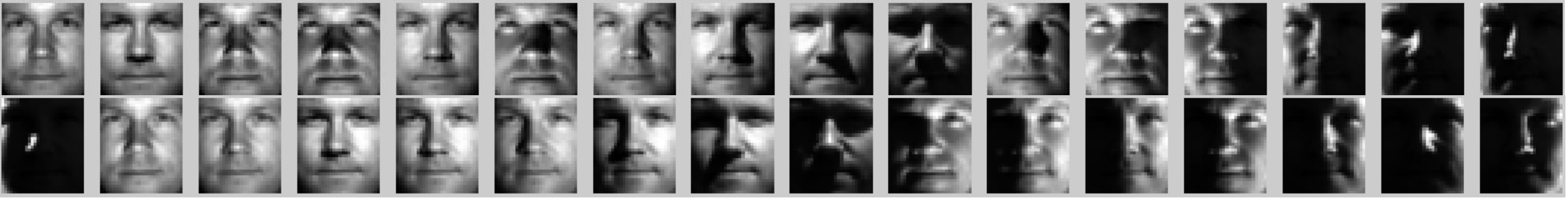} }
  \subfloat[Example subjects.] {\label{fig:yalesub} \includegraphics[angle=0, height=0.07\textwidth, width=.5\textwidth]{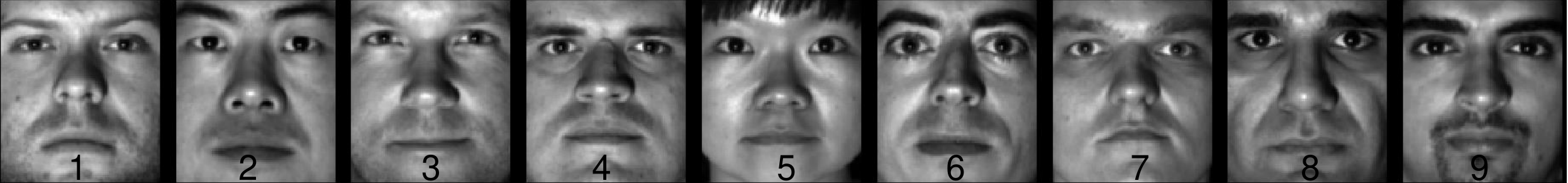}}
\caption{The Extended YaleB face dataset.}
\label{fig:yaledata}
%\end{center}
\end{figure*}

\begin{figure*} [ht]
\centering
 \subfloat[Subjects \{1, 2\}.] {\label{fig:subcode2_front} \includegraphics[angle=0, height=0.15\textwidth, width=.7\textwidth]{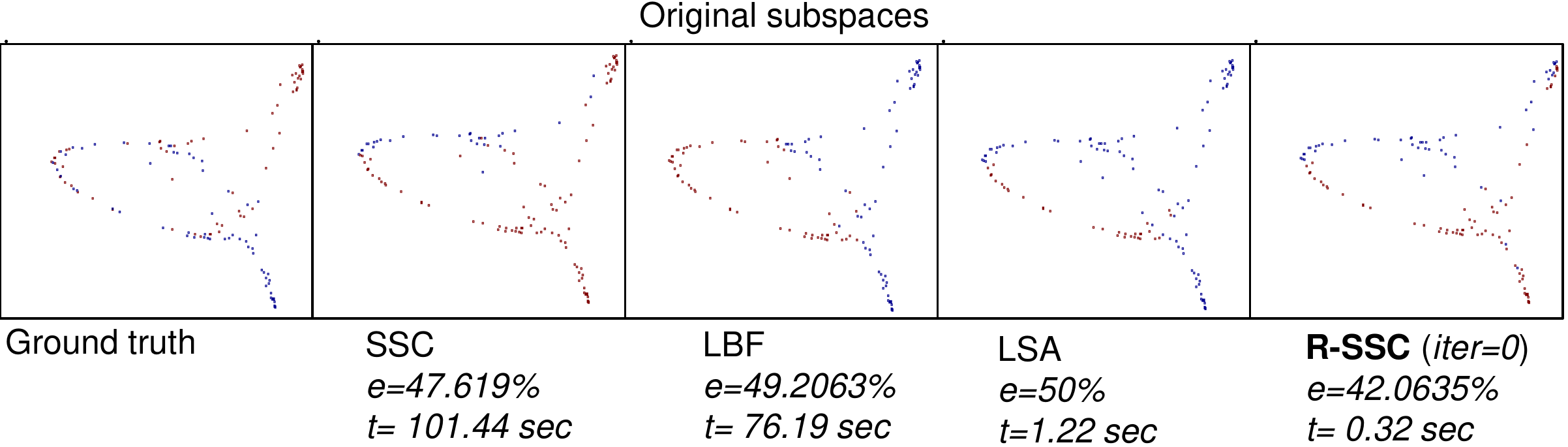} }
 \subfloat[Subjects \{1, 2\}.] {\label{fig:subcode2_side} \includegraphics[angle=0, height=0.15\textwidth, width=.3\textwidth]{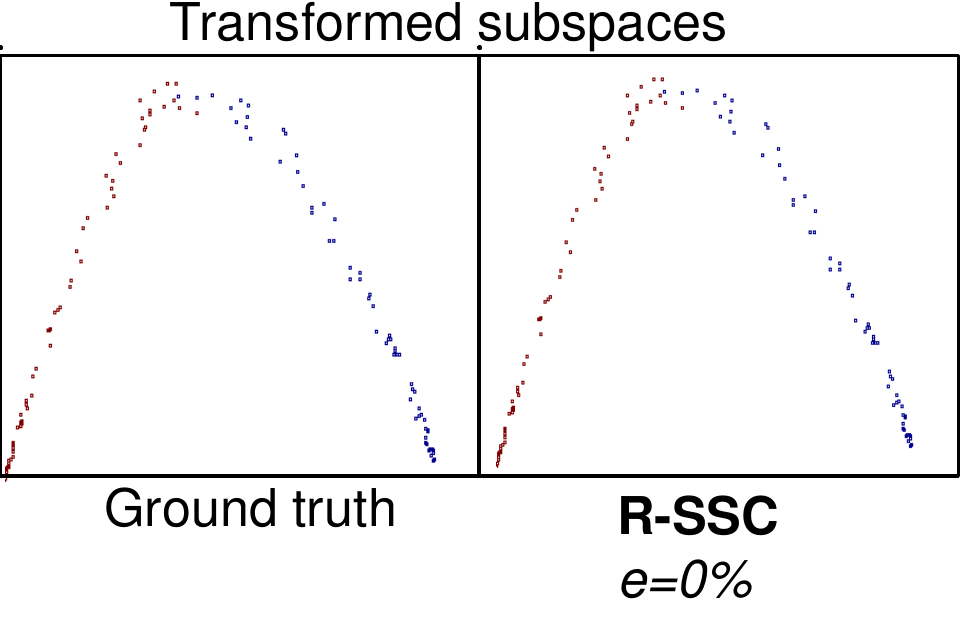}} \\
  \subfloat[Subjects \{2, 3\}.] {\label{fig:subcode2_profile} \includegraphics[angle=0, height=0.15\textwidth, width=.7\textwidth]{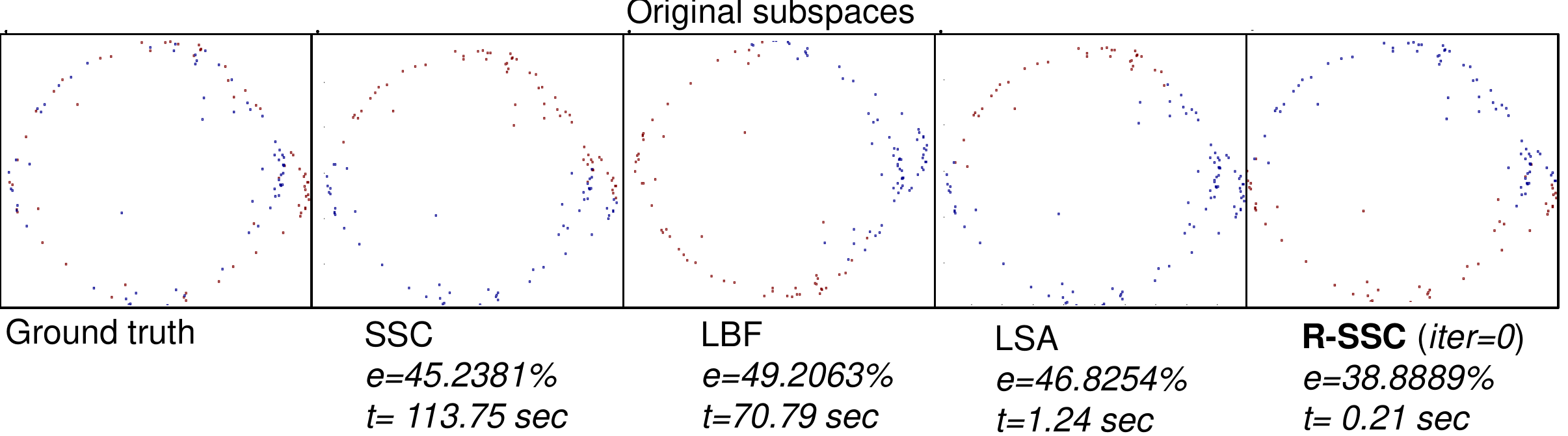}}
 \subfloat[Subjects \{2, 3\}.] {\label{fig:TF_subcode2_front} \includegraphics[angle=0, height=0.15\textwidth, width=.3\textwidth]{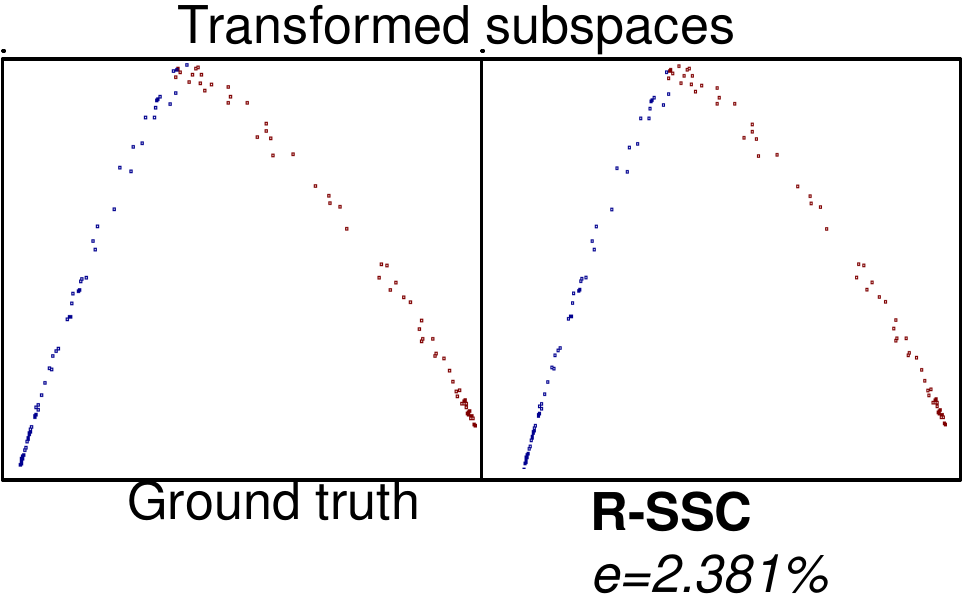} } \\
 \subfloat[Subjects \{4, 5, 6\}.] {\label{fig:TF_subcode2_side} \includegraphics[angle=0, height=0.15\textwidth, width=.7\textwidth]{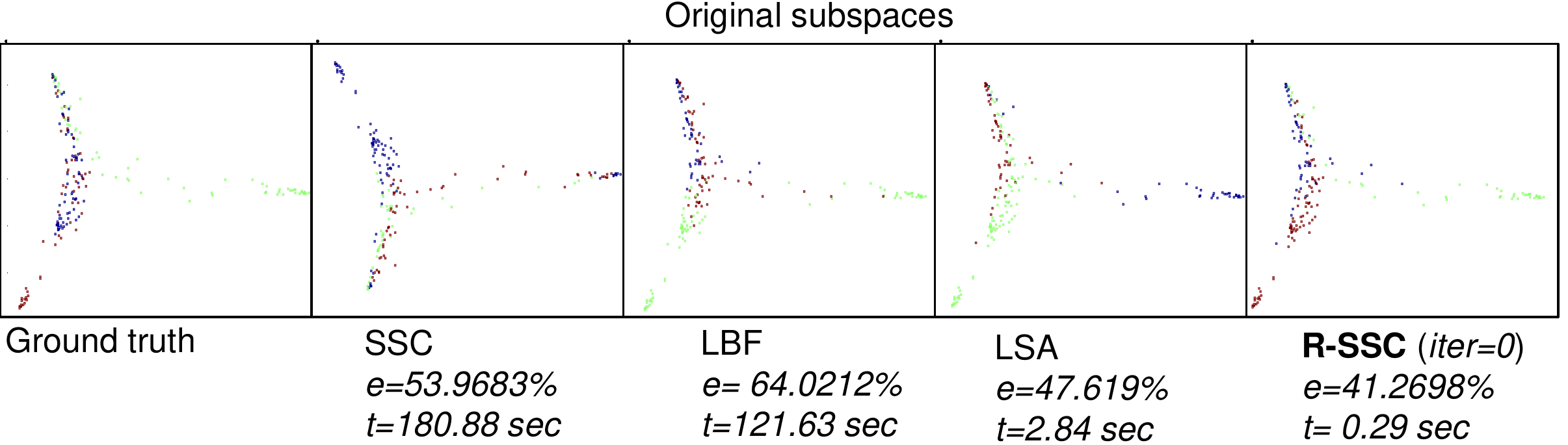}}
  \subfloat[Subjects \{4, 5, 6\}.] {\label{fig:TF_subcode2_profile} \includegraphics[angle=0, height=0.15\textwidth, width=.3\textwidth]{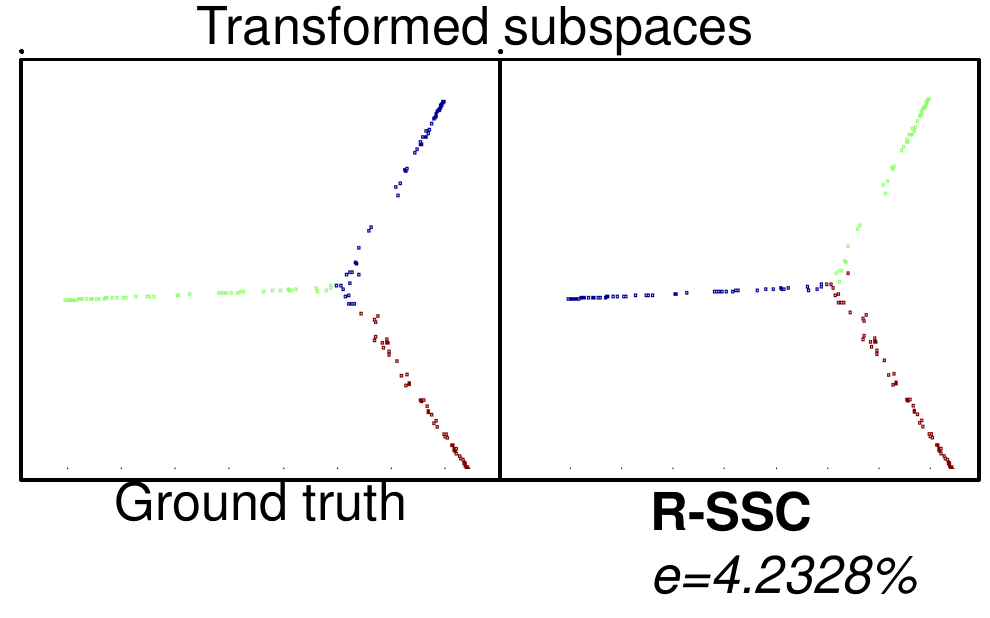}} \\
 \subfloat[Subjects \{7, 8, 9\}.] {\label{fig:TF_subcode2_side} \includegraphics[angle=0, height=0.15\textwidth, width=.7\textwidth]{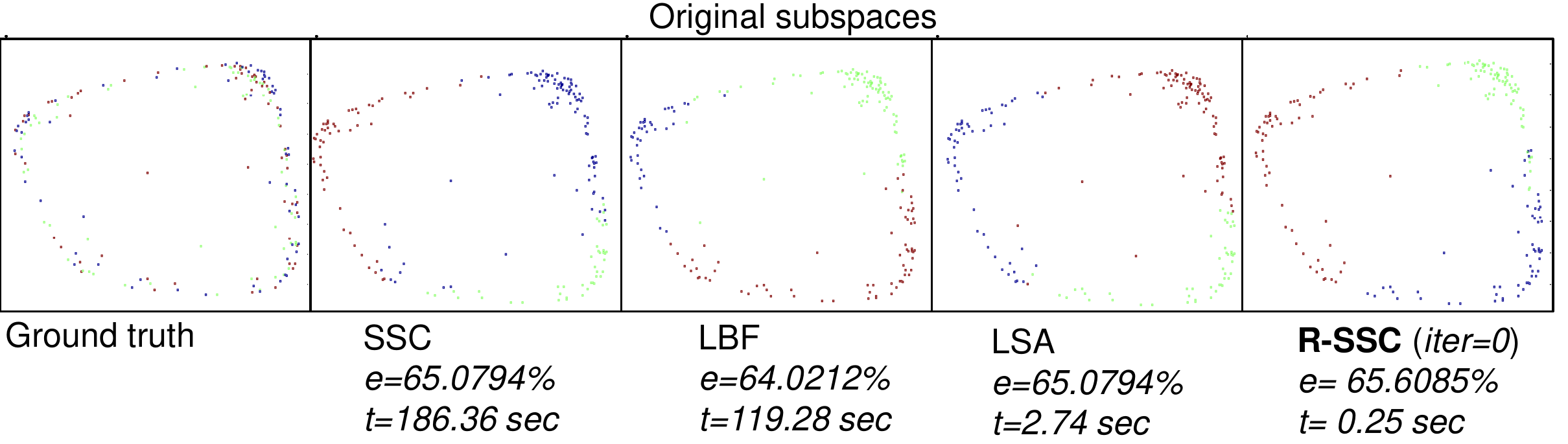}}
  \subfloat[Subjects \{7, 8, 9\}.] {\label{fig:TF_subcode2_profile} \includegraphics[angle=0, height=0.15\textwidth, width=.3\textwidth]{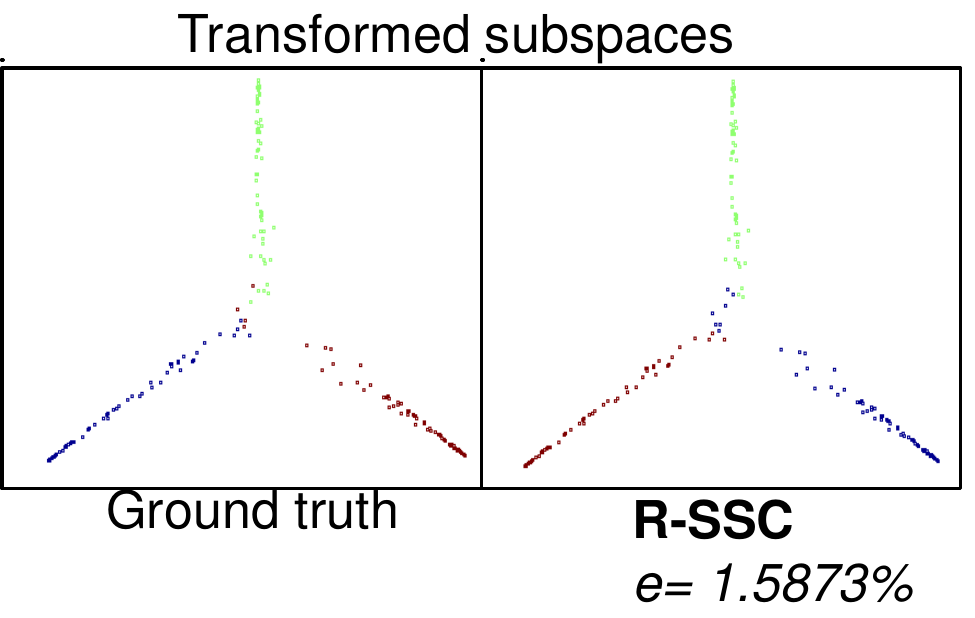}}\\
 \subfloat[All 9 subjects.] {\label{fig:TF_subcode2_side} \includegraphics[angle=0, height=0.15\textwidth, width=.7\textwidth]{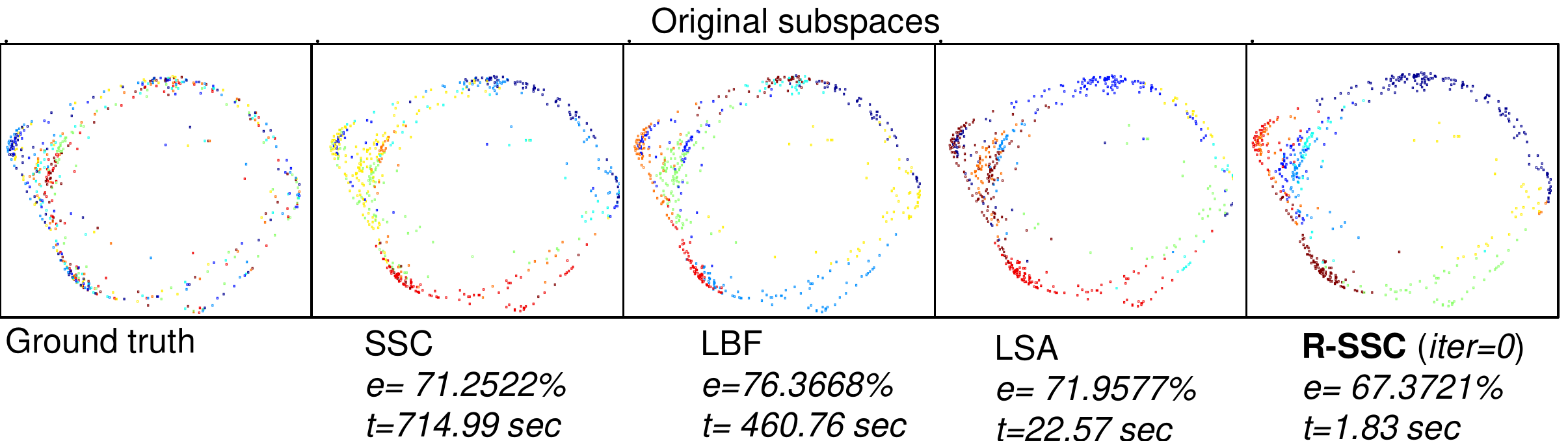}}
  \subfloat[All 9 subjects.] {\label{fig:TF_subcode2_profile} \includegraphics[angle=0, height=0.15\textwidth, width=.3\textwidth]{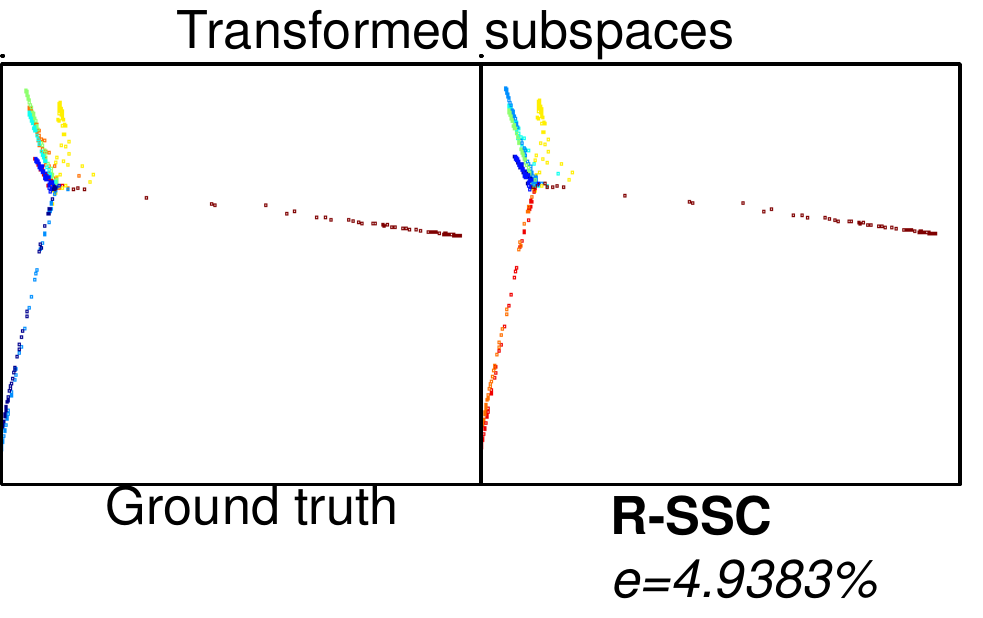}}
\caption{Misclassification rate (\emph{e})  and running time (\emph{t}) on clustering 2 subjects, 3 subjects and 9 subjects.
The proposed R-SSC outperforms state-of-the-art methods for both accuracy and running time.
With the proposed RSC framework, the clustering error of R-SSC is further reduced significantly, e.g., from $67.37 \%$ to $4.94 \%$ for the 9-subject case. Note how the classes are clustered in clean subspaces in the transformed domain (best viewed zooming on screen).
}
\label{fig:yaleacc}
%\end{center}
\end{figure*}

 In the Extended YaleB dataset, each of the 38 subjects is imaged under 64 lighting conditions, shown in Fig.~\ref{fig:yalelight}.
  We conduct the face clustering experiments on the first 9 subjects shown in Fig.~\ref{fig:yalesub}.
  We set the sparsity value $K=10$ for R-SSC, and perform $100$ iterations for the gradient descent updates while learning the transformation.
Fig.~\ref{fig:yaleacc} shows error rate (\emph{e}) and running time (\emph{t}) on clustering
 subspaces of 2 subjects, 3 subjects, and 9 subjects.
 The proposed R-SSC outperforms state-of-the-art methods for both accuracy and running time. Using the proposed RSC framework  (that is, learning the transform),
  the misclassification errors of R-SSC are further reduced significantly, for example, from $42.06 \%$ to $0 \%$ for subjects \{1,2\}, and from $67.37 \%$ to $4.94 \%$ for the 9 subjects.

\subsection{Application to Motion Segmentation}

\begin{figure}[!h]
\begin{minipage}[b]{0.5\linewidth}\centering
 \includegraphics[angle=0, height=0.20\textwidth, width=.95\textwidth]{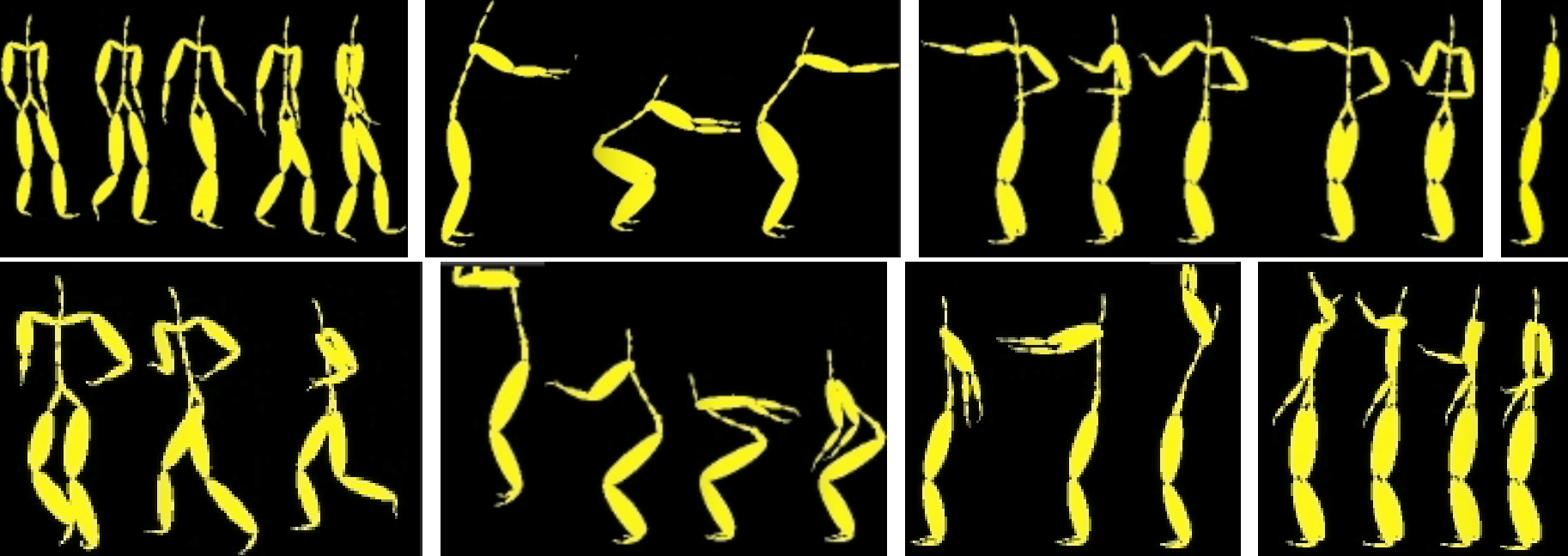}
\caption{Eight actions performed by subject 86 in the CMU motion capture dataset.}
\label{fig:mocap}
\end{minipage}
\begin{minipage}[b]{0.5\linewidth}\centering
{\scriptsize
	\begin{tabular}{|l|l|}
	\hline
Method & Misclassification (\%)  \\
	\hline
 \hline
SSC \cite{SSC} & 21.8693  \\
LSA \cite{LSA} & 17.8766  \\
LBF \cite{SLBF} & 33.8475  \\
\hline
\hline
R-SSC & 19.0653 \\
R-SSC+RSC & \textbf{3.902} \\
\hline
	\end{tabular}
}
\captionof{table}{Motion segmentation error.}
%\caption{Face recognition rate (\%) on the Extended YaleB face dataset.}
\label{tab:mocapacc}
\end{minipage}
\end{figure}

In the Mocap dataset, we consider the trial 2 sequence performed by subject 86, which consists of eight different actions shown in Fig.~\ref{fig:mocap}.
As discussed in \cite{rssc}, the data from each action lie in a low-dimensional subspace.
We achieve temporal motion segmentation by clustering sensor measurements corresponding to different actions.
We set the sparsity value $K=8$ for R-SSC and downsample the sequence by a factor 2 as \cite{rssc}.
As shown in Table~\ref{tab:mocapacc}, the proposed approach again significantly outperforms state-of-the-art clustering methods for motion segmentation.

\subsection{Discussion on the Size of the Transformation Matrix $\mathbf{T}$}

In the experiments presented above, images are resized to $16 \times 16$. Thus, the learned subspace transformation $\mathbf{T}$ is of size $256 \times 256$.
If we learn a transformation of size $r \times 256$ with $r<256$, we enable dimension reduction while performing subspace transformation. Through experiments, we notice that the peak clustering accuracy is usually obtained when $r$ is smaller than the dimension of the ambient space.
For example, in Fig.~\ref{fig:yaleacc}, through exhaustive search for the optimal $r$, we observe the misclassification rate reduced from $2.38\%$ to $0\%$ for subjects \{2, 3\} at $r=96$, and from $4.23\%$ to $0\%$ for subjects \{4, 5, 6\} at $r=40$.
As discussed before, this provides a framework to sense for clustering and classification, connecting the work here presented with the extensive literature on compressed sensing, and in particular for sensing design, e.g., \cite{CS1}.
We plan to study in detail the optimal size of the learned transformation matrix for subspace clustering and further investigate such connections with compressed sensing.

\section{Conclusion}

We presented a subspace low-rank transformation approach to robustify subspace clustering.
Using matrix rank as the optimization criteria, we learn a subspace transformation that reduces variations within the subspaces, and increases separations between the subspaces for more accurate subspace clustering.
We demonstrated that the proposed approach significantly outperforms state-of-the-art subspace clustering methods.

{\small
\bibliographystyle{ieee}
\bibliography{RSC}

\begin{thebibliography}{10}\itemsep=-1pt

\bibitem{9point}
R.~Basri and D.~W. Jacobs.
\newblock Lambertian reflectance and linear subspaces.
\newblock {\em IEEE Trans. on Patt. Anal. and Mach. Intell.}, 25(2):218--233,
  February 2003.

\bibitem{eigenmap}
M.~Belkin and P.~Niyogi.
\newblock Laplacian eigenmaps for dimensionality reduction and data
  representation.
\newblock {\em Neural Computation}, 15:1373--1396, 2003.

\bibitem{rpca}
E.~J. Cand\`{e}s, X.~Li, Y.~Ma, and J.~Wright.
\newblock Robust principal component analysis?
\newblock {\em J. ACM}, 58(3):11:1--11:37, June 2011.

\bibitem{CS1}
W.~R. Carson, M.~Chen, M.~R.~D. Rodrigues, R.~Calderbank, and L.~Carin.
\newblock Communications-inspired projection design with application to
  compressive sensing.
\newblock {\em SIAM J. Imaging Sci.}, 5(4):1185--1212, 2012.

\bibitem{scc}
G.~Chen and G.~Lerman.
\newblock Spectral curvature clustering ({SCC}).
\newblock {\em International Journal of Computer Vision}, 81(3):317--330, Mar.
  2009.

\bibitem{SSC}
E.~Elhamifar and R.~Vidal.
\newblock Sparse subspace clustering: Algorithm, theory, and applications.
\newblock {\em IEEE Trans. on Patt. Anal. and Mach. Intell.}, 2013.

\bibitem{yaleb}
A.~S. Georghiades, P.~N. Belhumeur, and D.~J. Kriegman.
\newblock From few to many: Illumination cone models for face recognition under
  variable lighting and pose.
\newblock {\em IEEE Trans. on Patt. Anal. and Mach. Intell.}, 23(6):643--660,
  June 2001.

\bibitem{llmc}
A.~Goh and R.~Vidal.
\newblock Segmenting motions of different types by unsupervised manifold
  clustering.
\newblock In {\em Proc. IEEE Computer Society Conf. on Computer Vision and
  Patt. Recn.}, 2007.

\bibitem{ocr}
T.~Hastie and P.~Y. Simard.
\newblock Metrics and models for handwritten character recognition.
\newblock {\em Statistical Science}, 13(1):54--65, 1998.

\bibitem{3dalign}
O.~Kuybeda, G.~A. Frank, A.~Bartesaghi, M.~Borgnia, S.~Subramaniam, and
  G.~Sapiro.
\newblock A collaborative framework for {3D} alignment and classification of
  heterogeneous subvolumes in cryo-electron tomography.
\newblock {\em Journal of Structural Biology}, 181:116–127, 2013.

\bibitem{robustsubspace}
G.~Liu, Z.~Lin, and Y.~Yu.
\newblock Robust subspace segmentation by low-rank representation.
\newblock In {\em International Conference on Machine Learning}, 2010.

\bibitem{spectral}
U.~Luxburg.
\newblock A tutorial on spectral clustering.
\newblock {\em Statistics and Computing}, 17(4):395--416, Dec. 2007.

\bibitem{alc}
Y.~Ma, H.~Derksen, W.~Hong, and J.~Wright.
\newblock Segmentation of multivariate mixed data via lossy data coding and
  compression.
\newblock {\em IEEE Trans. on Patt. Anal. and Mach. Intell.}, 29(9):1546--1562,
  2007.

\bibitem{RASL}
Y.~Peng, A.~Ganesh, J.~Wright, W.~Xu, and Y.~Ma.
\newblock {RASL}: Robust alignment by sparse and low-rank decomposition for
  linearly correlated images.
\newblock In {\em Proc. IEEE Computer Society Conf. on Computer Vision and
  Patt. Recn.}, 2010.

\bibitem{LLE}
S.~T. Roweis and L.~K. Saul.
\newblock Nonlinear dimensionality reduction by locally linear embedding.
\newblock {\em Science}, 290:2323--2326, 2000.

\bibitem{lrsalient}
X.~Shen and Y.~Wu.
\newblock A unified approach to salient object detection via low rank matrix
  recovery.
\newblock In {\em Proc. IEEE Computer Society Conf. on Computer Vision and
  Patt. Recn.}, June 2012.

\bibitem{ga-ssc}
M.~Soltanolkotabi and E.~J. Candes.
\newblock A geometric analysis of subspace clustering with outliers.
\newblock {\em The Annals of Statistics}, 40(4):2195--2238, 2012.

\bibitem{rssc}
M.~Soltanolkotabi, E.~Elhamifar, and E.~J. Cand{\`e}s.
\newblock Robust subspace clustering.
\newblock {\em CoRR}, abs/1301.2603, 2013.

\bibitem{pablo-lr}
P.~Sprechmann, A.~M. Bronstein, and G.~Sapiro.
\newblock Learning efficient sparse and low rank models.
\newblock {\em CoRR}, abs/1212.3631, 2012.

\bibitem{sfm}
C.~Tomasi and T.~Kanade.
\newblock Shape and motion from image streams under orthography: a
  factorization method.
\newblock {\em International Journal of Computer Vision}, 9:137--154, 1992.

\bibitem{SubspaceClustering}
R.~Vidal.
\newblock Subspace clustering.
\newblock {\em Signal Processing Magazine, IEEE}, 28(2):52--68, 2011.

\bibitem{gpca}
R.~Vidal, Y.~Ma, and S.~Sastry.
\newblock Generalized principal component analysis ({GPCA}).
\newblock In {\em Proc. IEEE Computer Society Conf. on Computer Vision and
  Patt. Recn.}, 2003.

\bibitem{llc}
J.~Wang, J.~Yang, K.~Yu, F.~Lv, T.~Huang, and Y.~Gong.
\newblock Locality-constrained linear coding for image classification.
\newblock In {\em Proc. IEEE Computer Society Conf. on Computer Vision and
  Patt. Recn.}, 2010.

\bibitem{nssc}
Y.~Wang and H.~Xu.
\newblock Noisy sparse subspace clustering.
\newblock In {\em International Conference on Machine Learning}, 2013.

\bibitem{subdifferential}
G.~A. Watson.
\newblock Characterization of the subdifferential of some matrix norms.
\newblock {\em Linear Algebra and Applications}, 170:1039--1053, 1992.

\bibitem{Wright09}
J.~Wright, M.~Yang, A.~Ganesh, S.~Sastry, and Y.~Ma.
\newblock Robust face recognition via sparse representation.
\newblock {\em IEEE Trans. on Patt. Anal. and Mach. Intell.}, 31(2):210--227,
  2009.

\bibitem{LSA}
J.~Yan and M.~Pollefeys.
\newblock A general framework for motion segmentation: independent,
  articulated, rigid, non-rigid, degenerate and non-degenerate.
\newblock In {\em Proc. European Conference on Computer Vision}, 2006.

\bibitem{SLBF}
T.~Zhang, A.~Szlam, Y.~Wang, and G.~Lerman.
\newblock Hybrid linear modeling via local best-fit flats.
\newblock {\em International Journal of Computer Vision}, 100(3):217--240,
  2012.

\bibitem{TILT}
Z.~Zhang, X.~Liang, A.~Ganesh, and Y.~Ma.
\newblock {TILT}: transform invariant low-rank textures.
\newblock In {\em Proceedings of the 10th Asian conference on Computer vision},
  2011.

\end{thebibliography}
}

\end{document}